\documentclass{article}
\usepackage{spconf,amsmath,graphicx}

\usepackage[capitalise, noabbrev]{cleveref}
\usepackage{booktabs}
\usepackage{comment}
\usepackage{multirow, tabularx}
\usepackage{array}
\usepackage{nth}
\usepackage{url}
\usepackage[table]{xcolor}

\newcolumntype{Y}{>{\centering\arraybackslash}X}

\title{DIALOGUE STRATEGY ADAPTATION TO NEW ACTION SETS USING MULTI-DIMENSIONAL MODELLING}
\name{Simon Keizer, Norbert Braunschweiler, Svetlana Stoyanchev, and Rama Doddipatla}  
\address{Cambridge Research Laboratory, Toshiba Europe Limited, United Kingdom}
\begin{document}
\ninept
\maketitle
\begin{abstract}
A major bottleneck for building statistical spoken dialogue systems for new domains and applications is the need for large amounts of training data.  To address this problem, we adopt the multi-dimensional approach to dialogue management and evaluate its potential for transfer learning.  Specifically, we exploit pre-trained task-independent policies to speed up training for an extended task-specific action set, in which the single summary action for requesting a slot is replaced by multiple slot-specific request actions.  Policy optimisation and evaluation experiments using an agenda-based user simulator show that with limited training data, much better performance levels can be achieved when using the proposed multi-dimensional adaptation method.  We confirm this improvement in a crowd-sourced human user evaluation of our spoken dialogue system, comparing partially trained policies.  The multi-dimensional system (with adaptation on limited training data in the target scenario) outperforms the one-dimensional baseline (without adaptation on the same amount of training data) by 7\% perceived success rate.
\end{abstract}
%


\section{Introduction}
\label{sec:intro}

One of the main challenges in spoken dialogue system development is their scalability to new domains and applications.  A statistical spoken dialogue system can be built efficiently for a large slot-filling domain when sufficient annotated data in that domain is available.  However, as domains are expanded with new slots, and new applications emerge that require new, task-specific system actions, in-domain annotated datasets are typically hard to get by.  Therefore, numerous efforts have been made to use transfer learning techniques \cite{pan_survey_2010, taylor_survey_2009} to efficiently develop dialogue systems for new domains with limited or no data in the target domain.  Most of the efforts in statistical dialogue management and action selection in particular focus on adaptation to newly introduced slots and values, with the underlying task unchanged \cite{wang-etal-2015-learning-domain, chen-etal-icassp2018, GASIC2017552}.

The multi-dimensional approach to dialogue modelling \cite{Bunt:2011et} offers the potential to exploit its principled separation of domain- and application-independent aspects of dialogue to adapt to new domains as well as new applications.  In \cite{keizer-etal-2019-user}, a multi-dimensional statistical dialogue manager was presented and it was demonstrated that policy optimisation for a target domain could be improved by re-using the policies for the application-independent dimensions (Social Obligations Management and Auto-feedback; see \cref{sec:mdim-dm}) that were pre-trained in a source domain.  
However, the addressed use-case was limited to adapting across very similar domains, where the set of task-specific actions was the same and only the set of slots and values changed.

In this paper, we adopt the multi-dimensional approach to dialogue management \cite{keizer-bunt-2006-multidimensional, keizer-rieser-semdial2017, keizer-etal-2019-user}, but rather than adapting to new slots and values, we focus on adapting to a different policy action set, which is typically required when developing a dialogue system for a new application.  In addition, we present two extensions of the model: 1) whereas \cite{keizer-etal-2019-user} restricted their model to allow only single dialogue acts being generated (despite the multi-agent design), we allow combinations of auto-feedback and task acts that support implicit confirmation, and 2) the agent that evaluates the generated dialogue act candidates from the different dimensions (see \cref{sec:mdim-dm}) is trained jointly with the other agents, rather than hand-coded.

Using the improved multi-dimensional design, we present policy optimisation and evaluation experiments using our simulated user and error model, demonstrating significant improvements in performance on limited training data when using the proposed adaptation method (\cref{sec:pol-opt}).  Furthermore, a human user evaluation was carried out to confirm this result in more realistic conditions; the experimental design, spoken dialogue system implementation, and evaluation results are described in \cref{sec:hum-eval}.  \Cref{sec:related} contains a more detailed discussion of related work in the area of transfer learning for dialogue management.  The paper is wrapped up in \cref{sec:conclusion} with conclusions and directions for future work.


\begin{table*}[htb]
\setlength\extrarowheight{5pt}
\begin{tabularx}{\linewidth}{llp{.2\linewidth}p{.2\linewidth}p{0.28\linewidth}}
\toprule
{\bf Task}     & {\bf Auto-Feedback}  & {\bf SOM}              & {\bf All}                  & \textit{\textbf{Example utterance}} \\
\midrule
{\sc offer}    &                      &                        & {\sc offer}                & \textit{How about the Rice Boat?} \\
{\sc offer}    & {\sc impl-confirm}   &                        & {\sc offer+impl-confirm}   & \textit{The Rice Boat is a nice Indian restaurant} \\
{\sc answer}   &                      &                        & {\sc answer}               & \textit{The address of the Rice Boat is \dots} \\
{\sc request}  &                      &                        & {\sc request}              & \textit{What price range did you have in mind?} \\
{\sc request}  & {\sc impl-confirm}   &                        & {\sc req+impl-confirm} & \textit{Okay, Indian food.  What price range did you have in mind?} \\
               & {\sc expl-confirm}   &                        & {\sc expl-confirm}         & \textit{You want Indian food, is that right?} \\
               & {\sc auto\_negative} &                        & {\sc auto\_negative}       & \textit{I did not quite catch that, could you please rephrase?} \\
               &                      & {\sc accept\_thanking} & {\sc accept\_thanking}     & \textit{You're welcome} \\
               &                      & {\sc return\_goodbye}  & {\sc return\_goodbye}      & \textit{Have a nice day} \\
{\sc none}     & {\sc none}           & {\sc none}             &                            & \\
\bottomrule
\end{tabularx}
\caption{The dialogue act agent action sets.  Note that {\sc impl-confirm} is essentially an inform act stating with a slot-value pair a preference that the system believes the user has, whereas {\sc expl-confirm} is essentially a propositional question asking the user whether they have a specific preference, expressed by a slot-value pair.}
\label{tab:actions}
\end{table*}

\section{Multi-dimensional dialogue management}\label{sec:mdim-dm}

\begin{figure}[htb]
\includegraphics[width=\linewidth]{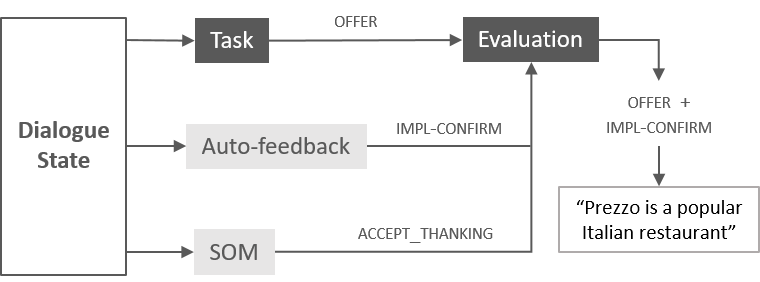}
\caption{Multi-agent action selection component, showing the Task, Auto-feedback, and SOM agents generating candidate dialogue acts, of which two are selected by the Evaluation agent.}
\label{fig:mdim-diagram}
\end{figure}

In conventional reinforcement learning-based statistical dialogue systems \cite{young-etal-pomdp-ieee2013}, the dialogue policy selects an action from a single set of possible actions in each turn.  In contrast to such `one-dimensional' systems, multi-dimensional dialogue systems employ multiple dialogue act agents, each dedicated to a different aspect of the dialogue process, using a policy that selects actions from its own specialised action set.  We propose and evaluate a model with three dimensions and corresponding agents: Auto-feedback, Social Obligation Management (SOM), and Task.  The Auto-feedback agent has actions for giving feedback to the user about processing their utterances, for example indicate non-understanding when the speech recogniser does not return any results (``I did not quite get that, could you please repeat?'') or provide articulate feedback in the form of an explicit/implicit confirmation when the system is unsure about the user's input (e.g., ``Expensive, you said?''), the SOM agent deals with greeting, thanking, apologising, and other social actions, and the Task agent focuses on the underlying task or activity (information navigation, tutoring, negotiation, etcetera).  Additional agents to support other dimensions could be introduced as well (e.g. for turn-taking and time-management \cite{ISO-SemAnnot_2012}), but the three agents currently included are considered to be the minimum requirement for a task-oriented multi-dimensional dialogue system.  An additional Evaluation agent is used for determining which of the generated dialogue act candidates are forwarded as final dialogue acts to the natural language generation component \cite{keizer-bunt-2007-evaluating}.  \Cref{fig:mdim-diagram} gives an outline of the multi-agent action selection component.

Besides reflecting the multi-dimensional nature of dialogue \cite{Bunt:2011et}, this design opens up opportunities for efficient adaptation of dialogue managers to new tasks and domains.  The Auto-feedback and SOM dimensions are considered to be domain and task independent, and therefore policies for these dimensions may be transferred, leaving only the Task policy to be trained from scratch in the target domain/application.

We introduce a new design of the multi-agent action selection model, in which the Evaluation agent is designed to allow two possible combinations of two dialogue acts.  Both of these combinations include an implicit confirmation from the Auto-feedback dimension, combined with either an offer (e.g., ``Prezzo is a popular Italian restaurant.''), or a request (e.g., ``You are looking for an Italian restaurant in which area?'') from the Task dimension.  \Cref{tab:actions} lists all allowed action combinations, organised along the three supported dimensions, as well as in a single action set for a one-dimensional baseline system.  In future versions, additional combinations can be considered, for example negative feedback combined with an apology (e.g., ``I'm sorry, I did not quite get that'').


\section{Policy optimisation and adaptation experiments}\label{sec:pol-opt}

The potential for transfer learning in a multi-dimensional dialogue manager has been demonstrated previously for the use-case of adapting from a hotel search application to a restaurant search application \cite{keizer-etal-2019-user}.  Here, we introduce a use-case where we stay within the restaurant domain, but adapt to a new action set.  Specifically, the action set of the Task agent is extended by replacing the summary action \textsc{request} for requesting a slot by separate request actions for each slot, e.g. \textsc{request-pricerange}.  Whereas in the source scenario the policy may select the \textsc{request} summary action, after which heuristics determine which slot is requested, in the target scenario the policy may select a request action for a particular slot, for example the area.  Therefore, in the target scenario with extended action set, the system learns automatically which slot to request, rather than relying on heuristics in the source scenario. 

In the DSTC-2 restaurant search domain \cite{henderson-etal-2014-second}, this means that the \textsc{request} summary action is replaced by the three slot-specific request actions, corresponding to the slots \texttt{food}, \texttt{area}, and \texttt{pricerange}.  Going back to the original action sets outlined in \cref{tab:actions}, the action set of the Task agent therefore grows from 4 (\textsc{offer}, \textsc{request}, \textsc{answer}, \textsc{none}) to 6 actions (\textsc{offer}, \textsc{request-food}, \textsc{request-area}, \textsc{request-pricerange}, \textsc{answer}, \textsc{none}), whereas the action set of the one-dimensional baseline (column `All') grows from 9 to 13 actions, since the extension applies to both \textsc{request} and \textsc{req+impl-confirm}.

Hence, we first train a multi-dimensional system for the source scenario, in which the Task agent uses a summary action for requesting slots ({\it mdim-src}).  This yields 4 trained policies, corresponding to the 3 conversational dimensions plus the evaluation agent.  Out of these, the policies for Auto-feedback and SOM are subsequently re-used and adapted when we train a system for the target scenario, in which the Task agent uses the extended action set ({\it mdim-ada}).  As baselines, we also train a multi-dimensional system for the target scenario without re-using the pre-trained Auto-feedback and SOM policies ({\it multi-dim}) and a one-dimensional system for the target scenario in which all action combinations from the multi-dimensional system are taken as single actions into one action set ({\it one-dim}).

%

\subsection{Training details}

All action selection models are trained in online interaction with an agenda-based user simulator \cite{schatzmann-etal-2007-agenda}.  In each case, all 4 policies are trained simultaneously using Monte Carlo Control reinforcement learning with linear value function approximation \cite{Sutton1998}.  Each policy selects actions from its own action set, based on the approximated values given the current state.  The values predict the long-term cumulative reward when taking an action in a given state and following the policy in subsequent turns, where the state is represented by a set of features extracted from the full dialogue state.  During training, the policies use Boltzmann exploration, i.e., actions are sampled from a softmax distribution applied to the estimated values.  The temperature hyper-parameter of the softmax is decayed linearly, gradually reducing the level of exploration until the policy only selects actions with the highest estimated value.  The weights of the linear value function are updated after every dialogue/episode, based on a shared reward signal.  The policies coordinate their actions only indirectly through the shared rewards, i.e., each policy operates independently without any direct communication with other policies.

The rewards are calculated at each turn, combining rewards obtained from the simulated user with internal rewards.  The user gives a reward of +100 upon task completion: in the restaurant search domain, this is when the system has recommended a restaurant matching the user's preferences and has provided all requested information about this restaurant.  Such `user goals' are randomly initialised from the domain ontology at the start of each dialogue, and fed to the simulated user.  In addition, the user issues a penalty when the system fails to respond to a thanking action or inserts a social act when it is not called for (-5 for each occurrence).  This is to force the system to learn basic reactive social behaviour, though we are aware that this might be experienced as repetitive and unnatural.  In future work, we will consider learning more sophisticated social patterns.  Internally, a penalty of -1 is applied for each turn (to encourage shorter dialogues) and a penalty of -25 when a `processing problem' is encountered and the system does not signal this to the user with a feedback act.  A processing problem is recorded in the dialogue state when speech recognition or natural language understanding fails, i.e., returns no results.  During training, such processing problems are simulated randomly in 5\% of the user turns, discarding the original simulated user act.

The Evaluation agent from \cite{keizer-etal-2019-user} was implemented via a set of rules, based on definitions from the dialogue act annotation standard \cite{ISO-SemAnnot_2012}.  Here, we implement it as another reinforcement learning agent, which takes as input the candidate dialogue act for each dimension and selects which dialogue act combination will be passed on to the natural language generation component.  Given that there are 3 supported dimensions, the actions correspond to the 8 possible combinations of dimensions that can be selected.  Using an action masking mechanism, we ensure that the Evaluation agent only allows single dialogue acts or a combination of \textsc{impl-confirm} (Auto-feedback dimension) with \textsc{offer} or with any of the three \textsc{request} actions (Task dimension).

\subsection{Results}

\begin{figure}[htb]
\includegraphics[width=\linewidth]{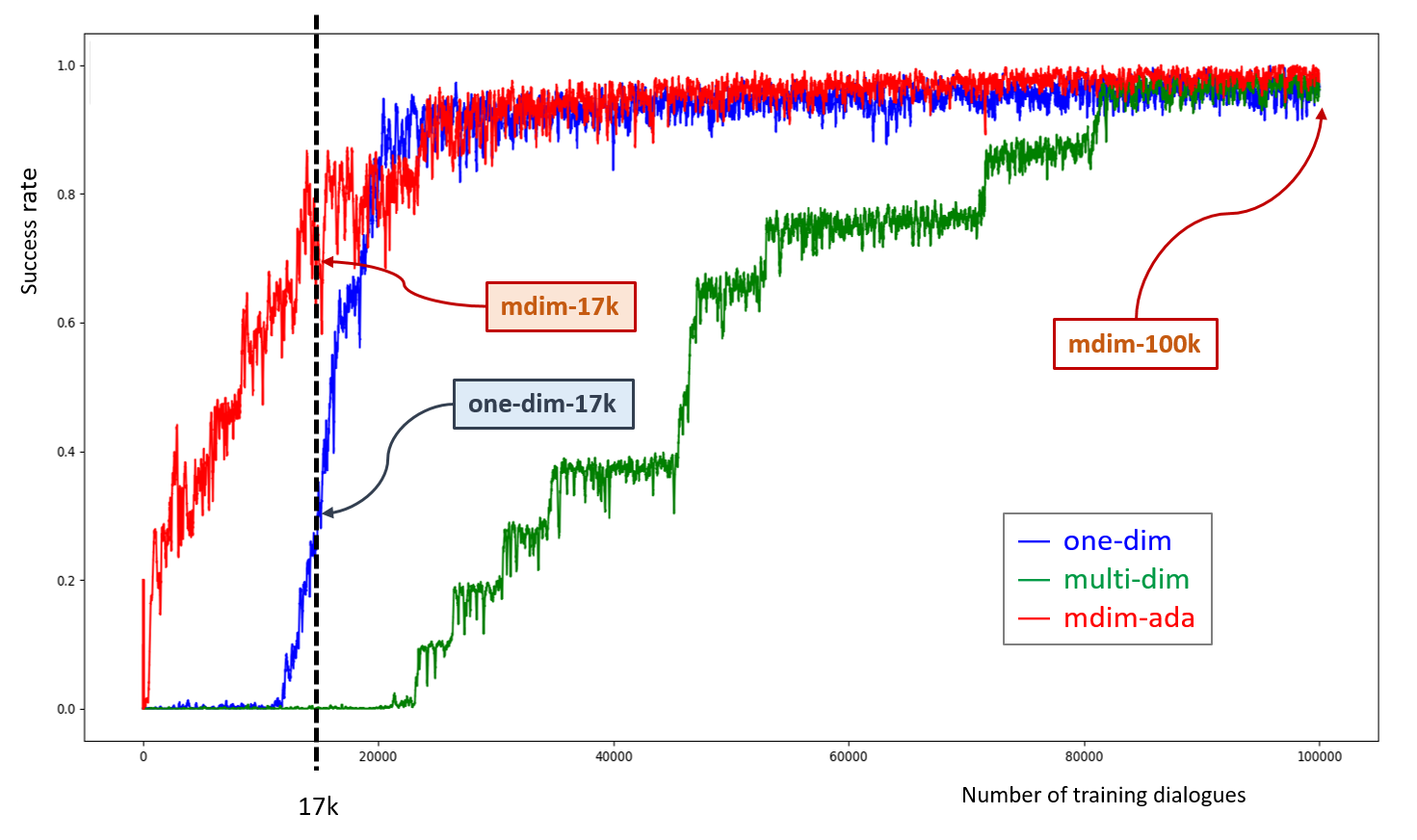}
\caption{Learning curves for the target scenario in terms of success rate for the three training methods.  Note that the results are averaged over 10 training runs in each setting; in each training run, success rates over a sliding window of 100 dialogues are recorded.}
\label{fig:sim-tra}
\end{figure}

The policy optimisation results in simulation for the target scenario are shown in \cref{fig:sim-tra}.  The learning curves (here shown in terms of success rates) quite clearly show that the performance levels are much higher in the early stages of training when using our proposed multi-dimensional adaptation method ({\it mdim-ada}, in red) than when training a multi-dimensional action selection model from scratch ({\it multi-dim}, in green).  In the first 20k training dialogues we also observe much higher success rates compared to the one-dimensional baseline ({\it one-dim}, in blue).

After training all policies, we ran evaluations of the fully trained policies over 3000 simulated dialogues each, at 25\% semantic error rate.  The first three rows in \cref{tab:sim-eval} correspond to the systems with the extended action set (the target scenario) that were also shown in \cref{fig:sim-tra}.  These systems all score similar average rewards, although the adapted system (mdim-ada) gets a slightly higher success rate, but with slightly longer dialogues.  Since the system with the original action set (the source scenario) covers the same range of full system dialogue acts (through its summary actions and heuristics for mapping them to complete dialogue acts), it has the same functionality as the extended system, and can therefore be included in the evaluation for comparison.  As this system ({\it mdim-src}) achieves similar scores, automatically learning request actions for each slot did not result in improved performance levels in this particular scenario.  Whether improvements can be achieved in this way, however, depends on the nature of the domain ontology and database, and on the behaviour of the users.

\begin{table}[htb]
\begin{tabularx}{\linewidth}{lXXXXXX}
\toprule
{\bf System} & {\bf ASet} & {\bf ADA} & {\bf NLU} & {\bf Succ} & {\bf Len} & {\bf Rew} \\
\midrule
one-dim        & ext & no  & sim & 96.0 & 9.28 & 86.06  \\
multi-dim      & ext & no  & sim & 96.7 & 9.49 & 85.49  \\
mdim-ada       & ext & yes & sim & 98.3 & 9.71 & 87.92  \\
mdim-src       & sum & no  & sim & 97.2 & 9.57 & 87.15  \\
one-dim-asu    & ext & no  & asu & 99.5 & 7.69 & 91.83  \\
mdim-ada-asu   & ext & yes & asu & 88.9 & 7.51 & 81.41  \\
%
\bottomrule
\end{tabularx}
\caption{Evaluation of fully trained policies over 3000 simulated dialogues. ASet: Action Set (\textbf{sum}mary/\textbf{ext}ended), ADA: adapted (yes/no), NLU: Natural Language Understanding (\textbf{asu}/\textbf{sim}ulated), Succ: Success Rate, Len: Average Length, Rew: Average Reward}
\label{tab:sim-eval}
\end{table}

\begin{table*}[htb]
\begin{tabularx}{\linewidth}{Y|rrr|rrr|rrr|Y}
\toprule
\multicolumn{1}{c}{\bf System} & \multicolumn{3}{c|}{\bf Success rate} & \multicolumn{3}{c|}{\bf Average length} & \multicolumn{3}{c|}{\bf Average reward} & \multicolumn{1}{c}{\bf Num. dial's} \\
\midrule
one-dim    & 96.0 & 96.0 & 99.5  & 9.28 & 9.28 &  7.69  & 86.06 & 86.06 & 91.83  &     \\
\rowcolor[gray]{.8}
mdim-ada   & 98.3 & 98.3 & 88.9  & 9.71 & 9.71 &  7.51  & 87.92 & 87.92 & 81.41  & \multirow{-2}*{100k}        \\
\midrule
one-dim    & 59.1 & 61.7 & 64.9  & 8.02 & 11.64 & 10.15  & 49.05 & 45.78 & 52.13  & 18k     \\
mdim-ada   & 73.5 & 73.8 & 73.3  & 8.95 & 10.01 &  7.71  & 62.18 & 60.77 & 64.25  &         \\
\midrule
\rowcolor[gray]{.8}
one-dim    & 48.9 & 57.5 & \textit{\textbf{60.3}} & 7.55 & 11.98 & \textit{\textbf{10.63}}  & 37.31 & 38.05 & \textit{\textbf{45.82}}  & \textit{\textbf{17k}}     \\
\rowcolor[gray]{.8}
mdim-ada   & 77.1 & 79.1 & \textit{\textbf{78.5}}  & 9.49 & 10.70 & \textit{\textbf{7.75}}  & 65.55 & 64.99 & \textit{\textbf{68.72}}  &         \\
\midrule
one-dim    & 45.3 & 48.5 & 54.1  & 7.30 & 11.60 & 10.16  & 34.07 & 30.04 & 39.77  & 16k     \\
mdim-ada   & 83.5 & 83.5 & 83.2  & 9.32 & 10.36 &  8.13  & 73.04 & 71.18 & 74.21  &         \\
\midrule
one-dim    & 30.6 & 33.4 & 37.5  & 5.61 & 11.94 & 10.70  & 19.86 & 13.95 & 21.48  & 15k     \\
mdim-ada   & 64.5 & 64.3 & 62.3  & 9.26 & 10.14 &  8.08  & 52.67 & 50.54 & 52.44  &         \\
\midrule
           & {\bf no-expl} & {\bf \;\;expl} & {\bf expl+asu}  & {\bf no-expl} & {\bf \;\;expl} & {\bf expl+asu}  & {\bf no-expl} & {\bf \;\;expl} & {\bf expl+asu} & \\
\bottomrule
\end{tabularx}
\caption{Evaluation of partially trained policies over 3000 simulated dialogues, reporting success rate, average dialogue length, and average reward in three settings.}
\label{tab:sim-eval-part}
\end{table*}

The results in the first 4 rows have been obtained using a rule-based state tracker that takes (simulated) user dialogue act hypotheses as input, as is the case during policy optimisation.  For the human user evaluation, we use the Action State Update (ASU) dialogue state tracker \cite{stoyanchev-etal-icassp2021} that takes user utterances as input (see also \cref{sec:sds}). 
We have therefore also evaluated the one-dim and mdim-ada policies with this tracker in the loop, and included a hybrid retrieval/template based natural language generation component that maps simulated user dialogue acts to natural language utterances that can be fed to the ASU component\footnote{Training the policies together with the ASU tracker is currently not practical, because its BERT model is too slow for running the required volume of training dialogues.}.  For the one-dimensional system, this results in higher scores ({\it one-dim-asu}), whereas for the adapted multi-dimensional system, we get considerably lower scores ({\it mdim-ada-asu}).  In the latter case, we note however that this is mainly due to one of the 10 policy combinations performing very badly: the policy repeatedly responds with negative feedback, after which the simulated user loses its patience and hangs up, resulting in 0\% success rate and -5.02 average reward.  In contrast, the other policies all achieve 98\% success rate.

In order to decide which partially trained policies to include in the human user evaluation, we have run simulated evaluations for the policies that were obtained after 15k, 16k, 17k, and 18k training dialogues.  This pre-selection is based on the learning curves in \cref{fig:sim-tra} and in particular the area where the success rate of the one-dimensional system is starting to build up, but is still strongly outperformed by the adapted multi-dimensional system.  The results in \cref{tab:sim-eval-part} report success rates, average dialogue lengths (in terms of user turns), and average rewards over 3000 simulated dialogues at 25\% error rate.  Each evaluation is carried out in three different settings, indicated in the bottom row of the table: 1) using the rule-based state tracker and no policy exploration, i.e. the policies always select the action with the highest estimated value ({\it no-expl}); 2) using the rule based state tracker and policy exploration at the level determined by the temperature setting at the corresponding stage of training ({\it expl}); and 3) using the ASU state tracker and policy exploration ({\it expl+asu}).

At a relatively early training stage, the policy is still exploring the state-action space, and has not experienced many successful dialogues yet.  Therefore, the reported performance levels can differ between the {\it no-expl} and {\it expl} settings.  Overall, the results in \cref{tab:sim-eval-part} show lower average rewards in the {\it expl} setting, but higher success rates.
Furthermore, the performance levels are much higher when evaluating with the ASU tracker in the loop.  The likely reason for this is that no semantic error model was used in this setting, i.e., the true user dialogue acts from the simulator were used to generate the input user utterances for the tracker.  Hence, the language understanding performance is probably much better in this setting, and therefore the action selection performance as well.

Based on these results, we selected the one- and multi-dimensional models that were trained on 17k dialogues for the human evaluation, as well as the fully-trained multi-dimensional system.  The three selected variants are highlighted in grey in \cref{tab:sim-eval-part} and also pointed out in \cref{fig:sim-tra}.


\begin{table*}[h]
\begin{tabularx}{\linewidth}{lcccccccc}
\toprule
{\bf System}  & {\bf Num.} & {\bf Average} & {\bf Q1 [\%]} & {\bf Q2 [\%]} & {\bf Q3 [1-6]} & {\bf Q4 [1-6]} & {\bf Q5 [1-6]} & {\bf Q6 [1-6]} \\
  & {\bf Dialogues} & {\bf Length} & {\bf Found Venue} &  {\bf DialSuccess} & {\bf Understand} & {\bf Recognise} & {\bf SysResponse} & {\bf Naturalness} \\
\midrule
one-dim-17k   & 203 & 7.69 (4.34) & 65.52 (3.34) & 63.05 (3.40) & 3.34 (1.68) & 3.46 (1.66) & 3.21 (1.72) & 3.01 (1.77) \\
mdim-ada-17k  & 201 & 6.52 (3.24) & 73.00 (3.15) & 70.00 (3.25) & 3.43 (1.58) & 3.56 (1.55) & 3.48 (1.59) & 3.35 (1.67) \\
mdim-ada-100k & 199 & 5.88 (2.38) & 78.89 (2.90) & 79.90 (2.85) & 3.76 (1.56) & 3.86 (1.50) & 3.93 (1.46) & 3.71 (1.62) \\
\bottomrule
\end{tabularx}
\caption{Human user evaluation results, where N is the number of dialogues, AvgLen is the average number of turns per dialogue, and Q1-6 are the scores obtained from the questionnaire. The scores for Q1-2 are percentages, and the standard deviation for each score is indicated in brackets.}
\label{tab:usr-eval}
\end{table*}

\section{Human User Evaluation}\label{sec:hum-eval}

Using the agenda-based user simulator and error model, we have shown that significant performance gains with limited training data can be achieved with the proposed adaptation method.  However, it remains to be seen to what extent these results are representative for the scenario of real users interacting with the system \cite{schatzmann-etal-2009-agenda, kreyssig-etal-2019}.  Ideally, the policy optimisation experiments should be carried out in online interaction with human users, which has been attempted with some success \cite{gasic-etal-asru2011, gasic-etal-icassp2013}.  However, this requires moving to more sample efficient optimisation methods and addressing many other technical challenges, which we will leave for future work.  Instead, we have run a human user evaluation in which we compared three system variants, and in particular two sets of partially trained policies, corresponding to the adapted multi-dimensional system and the non-adapted one-dimensional baseline, both trained on 17k dialogues only.

Users were recruited through the Amazon Mechanical Turk (AMT) crowd-sourcing platform, where they were given a task description (e.g., ``You are looking for a moderately priced French restaurant.  Make sure you get the phone number and address.'') and a link to the web-based interface that we have created.  To enable spoken dialogue on the web interface, we use the Google Web Speech API for both ASR (for recognising the user's speech) and TTS (for synthesizing the system's speech); our dialogue system server receives recognised user utterances and responds with system utterances, both in text form.

After finishing their conversation with the system, the user can `hang up' by pressing a button and receive a token which they can use to proceed on the AMT page, where they are given a questionnaire to fill out and submit, upon which they complete the task.  In the questionnaire, the subject is asked to state their opinion on 6 statements about the conversation, in the form of either a binary Yes or No (Q1 and Q2), or on a 6 point Likert scale (Q3 to Q6), ranging from `Strongly disagree' to `Strongly agree'.

\begin{itemize}
\setlength\itemsep{0.4em}
\item[Q1:] The system recommended a restaurant that matched my constraints. (Yes/No)
\item[Q2:] I got all the information I was looking for. (Yes/No)
\item[Q3:] The system understood what I was saying. (6 point Likert)
\item[Q4:] The system recognised my speech well. (6 point Likert)
\item[Q5:] The system's responses were appropriate. (6 point Likert)
\item[Q6:] The conversation felt natural. (6 point Likert)
\end{itemize}

\subsection{Spoken dialogue system implementation}\label{sec:sds}

Our dialogue system takes as input the user utterance text, in the form of the ASR top hypothesis obtained from the web interface.  A domain-independent dialogue act tagger is used to recognise social acts like goodbye and thanks, after which the utterance is passed to the Action State Update model for dialogue state tracking \cite{stoyanchev-etal-icassp2021}.  The dialogue state contains beliefs about the user goal (in terms of slot-value pairs and their confidence score), which slots are believed to be requested, which database items are or have been discussed, whether any processing problems have occurred (i.e. speech recognition or language understanding failed), and the dialogue act tags of previous user utterances.  Based on the dialogue state, one or more system dialogue acts are selected using one of the trained action selection models; the three models compared in the human evaluation are highlighted in \cref{tab:sim-eval-part}.  A set of templates is used to generate a natural language system response from the selected dialogue acts.  The response is synthesised on the web interface. 

In order to obtain results that are representative for each system variant, all 10 policies that were trained for each variant have been evaluated and the average results reported.  For each dialogue, one policy is selected from the pool of 10 using a round-robin system.

\subsection{Results}

The evaluation results are shown in \cref{tab:usr-eval}, including the (objective) average dialogue length and the (subjective) scores from the questionnaire.  As expected, the fully trained system mdim-ada-100k gets the best scores.  More importantly, across all metrics the partially trained mdim-ada-17k gets better scores than the partially trained baseline one-dim-17k.  Especially in terms of average dialogue length and the perceived partial and full task completion rates (Q1 resp. Q2) the difference between one-dim-17k and mdim-ada-17k is substantial, as was predicted by the experiments with the user simulator.  The scores for perceived understanding (Q3) and speech recognition (Q4) could somewhat explain these differences (rather than attributing them to the used policies), but their variance is quite large.  When assessing how appropriate the system responses appear to users (Q5), the difference between one-dimensional and multi-dimensional adaptation is larger.

Compared to the simulated evaluation results in \cref{tab:sim-eval-part}, the perceived success rates (Q2) are much lower in the case of the fully trained system (79.9\% vs. 88.9\%) and the multi-dimensional partially trained system (70\% vs. 78.5\%), whereas they are slightly higher in the case of the partially trained one-dimensional system (63.05\% vs. 60.03\%).  

In terms of average dialogue length, the human evaluation dialogues turn out to be shorter overall.  This suggests that the human subjects might have given up more quickly when the system failed to recommend a restaurant that met their constraints, i.e. the unsuccessful dialogues were shorter.  In contrast, the simulator keeps trying to complete the goal until the maximum number of user turns is reached (set to 30 turns in the configuration) or the system response act is repeated too many times (set to 3 times in the configuration).



\section{Related work}\label{sec:related}

In the area of transfer learning methods for dialogue management, most approaches have focused on cross-domain adaptation or multi-domain optimisation for slot-filling dialogue, where each domain is defined in terms of slots and their possible values.


%

In \cite{wang-etal-2015-learning-domain}, Domain-Independent Parameterisation (DIP) of dialogue state and action representations is introduced to enable transfer across domains. DIP seeks to train a dialogue policy that abstracts away from the specific slots and values in a particular domain, so it is applicable to the conversational search task in other domains as well.  The effectiveness of their method was demonstrated using an agenda-based user simulator when adapting from restaurant search to laptop search.  Furthermore, they carried out a human user evaluation, showing that their best transferred DIP policy performed at the same level as a non-transferred in-domain policy.

A Bayesian approach to dialogue management is described in \cite{GASIC2017552}, which uses Gaussian Process (GP) reinforcement learning, exploiting model priors of a generic dialogue policy for fast domain adaptation.  They also discuss a Bayesian committee machine approach, in which each domain is handled by a separate GP policy, but when only limited data is available for a specific domain, its policy may rely on the output of the other policies.  This approach has been further extended into a multi-agent learning setting, further improving performance levels during training with human users.

The Multi-Agent Dialogue Policy (MADP) is proposed in \cite{chen-etal-icassp2018}, which consists of several slot-specific agents and a slot-independent agent.  Adopting the Deep Q-Network (DQN) reinforcement learning framework \cite{Mnih-etal-nature}, the parameters of the slot-independent agent and the shared parameters of the slot-specific agents that have been learned for the source domain can be transferred to a new target domain.  So in this case, the multi-agent design is based on the slots in the domain definition.  The benefit of this method has only been demonstrated in simulated experiments, and is focused on adapting to new slots within the same restaurant/tourist information task.  A more recent incarnation of this approach is the AgentGraph framework \cite{Chen_etal-TASLP2019}, which employs Graph Neural Networks.  This method was evaluated using the PyDial benchmark, demonstrating successful transfer between restaurant search and laptop shopping tasks.

Where the agents in \cite{GASIC2017552} are associated with domains, and the agents in \cite{chen-etal-icassp2018, Chen_etal-TASLP2019} are associated with slots (except for the generic slot-independent agent), the agents in the multi-dimensional approach to dialogue management are associated with dimensions \cite{bunt-2006-dimensions}.  Although this approach is very different in nature, it is not necessarily incompatible with these other multi-agent approaches.  For example, it could benefit from the MADP approach by dividing up the Auto-feedback agent into sub-agents for each slot in the domain definition.

The multi-dimensional design presented in this paper is an extension of the previous models described in \cite{keizer-rieser-semdial2017, keizer-etal-2019-user}.  First, the system supports the generation of multiple dialogue acts in a single response; second, the Evaluation agent that was extended for this purpose is trained jointly with the other agents.  Furthermore, the adaptation use-case in previous work was limited to domain adaptation only, where the slots and values changed between source and target scenario, but not the action sets.  Finally, the human evaluation described in \cite{keizer-etal-2019-user} included fully trained policies only, showing that a multi-dimensional system could be trained to a performance level equivalent to a one-dimensional baseline.  In this paper, we have presented a human user evaluation to confirm the performance boost of partially adapted policies seen on simulated data when using the multi-dimensional adaptation method.






\section{Conclusion}\label{sec:conclusion}

We have presented a multi-dimensional approach to dialogue management, in which system response actions are selected through a combination of 4 dialogue policies, 2 of which are task-independent, and are therefore suitable for re-use when moving to a new application.  In simulated policy optimisation and evaluation experiments we have shown that by re-using and adapting these task-independent policies, significant performance gains can be achieved in the early stages of training.  As a first task adaptation use-case, we have looked at extending the task-specific action set with multiple slot-specific request actions, replacing the original summary request action that relied on heuristics to determine the slot to be requested.  To confirm the adaptation results obtained with an agenda-based user simulator, we have carried out a crowd-sourced human user evaluation.  
When trained on the same limited amount of training data, the proposed multi-dimensional adaptation strategy achieved significantly better results than the one-dimensional baseline (trained from scratch without any adaptation) on all subjective metrics, including task success and appropriateness of system responses, as well as on dialogue length.


In future work, we will consider more challenging use cases for applying the multi-dimensional adaptation method.  We will extend the role of the social agent to improve the naturalness of the interactions, explore methods for combining different types of agents (e.g., rule-based agents, or agents trained with supervised learning), and investigate the relation between multi-dimensional dialogue management and natural language generation.  Finally, to enable online learning with human users and therefore evaluate the proposed adaptation method more directly, we will explore more efficient reinforcement learning algorithms.

\bibliographystyle{IEEEbib}
\bibliography{simon_asru2021}

\begin{thebibliography}{10}

\bibitem{pan_survey_2010}
S.~J. Pan and Q~Yang,
\newblock ``A survey on transfer learning,''
\newblock {\em IEEE Transactions on Knowledge and Data Engineering}, vol. 22,
  no. 10, pp. 1345--1359, 2010.

\bibitem{taylor_survey_2009}
Matthew~E Taylor and Peter Stone,
\newblock ``Transfer learning for reinforcement learning domains: A survey,''
\newblock {\em The Journal of Machine Learning Research}, vol. 10, no. Jul, pp.
  1633--1985, 2009.

\bibitem{wang-etal-2015-learning-domain}
Zhuoran Wang, Tsung-Hsien Wen, Pei-Hao Su, and Yannis Stylianou,
\newblock ``Learning domain-independent dialogue policies via ontology
  parameterisation,''
\newblock in {\em Proceedings of the \nth{16} Annual Meeting of the Special
  Interest Group on Discourse and Dialogue}, Prague, Czech Republic, Sept.
  2015, pp. 412--416, Association for Computational Linguistics.

\bibitem{chen-etal-icassp2018}
Lu~Chen, Cheng Chang, Zhi Chen, Bowen Tan, Milica Gašić, and Kai Yu,
\newblock ``Policy adaptation for deep reinforcement learning-based dialogue
  management,''
\newblock in {\em 2018 IEEE International Conference on Acoustics, Speech and
  Signal Processing (ICASSP)}, 2018, pp. 6074--6078.

\bibitem{GASIC2017552}
Milica Gašić, Nikola Mrkšić, Lina~M. Rojas-Barahona, Pei-Hao Su, Stefan
  Ultes, David Vandyke, Tsung-Hsien Wen, and Steve Young,
\newblock ``Dialogue manager domain adaptation using gaussian process
  reinforcement learning,''
\newblock {\em Computer Speech \& Language}, vol. 45, pp. 552--569, 2017.

\bibitem{Bunt:2011et}
Harry Bunt,
\newblock ``{Multifunctionality in dialogue},''
\newblock {\em Computer Speech {\&} Language}, vol. 25, no. 2, pp. 222--245,
  2011.

\bibitem{keizer-etal-2019-user}
Simon Keizer, Ond{\v{r}}ej Du{\v{s}}ek, Xingkun Liu, and Verena Rieser,
\newblock ``User evaluation of a multi-dimensional statistical dialogue
  system,''
\newblock in {\em Proceedings of the \nth{20} Annual SIGdial Meeting on
  Discourse and Dialogue}, Stockholm, Sweden, Sept. 2019, pp. 392--398,
  Association for Computational Linguistics.

\bibitem{keizer-bunt-2006-multidimensional}
Simon Keizer and Harry Bunt,
\newblock ``Multidimensional dialogue management,''
\newblock in {\em Proceedings of the \nth{7} {SIG}dial Workshop on Discourse
  and Dialogue}, Sydney, Australia, July 2006, pp. 37--45, Association for
  Computational Linguistics.

\bibitem{keizer-rieser-semdial2017}
Simon Keizer and Verena Rieser,
\newblock ``Towards learning transferable conversational skills using
  multi-dimensional dialogue modelling,''
\newblock in {\em Proceedings \nth{21} Workshop on the Semantics and Pragmatics
  of Dialogue (SemDial/SaarDial)}, Saarbruecken, Germany, 2017.

\bibitem{young-etal-pomdp-ieee2013}
Steve Young, Milica Gašić, Blaise Thomson, and Jason~D. Williams,
\newblock ``{POMDP}-based statistical spoken dialog systems: A review,''
\newblock {\em Proceedings of the IEEE}, vol. 101, no. 5, pp. 1160--1179, 2013.

\bibitem{ISO-SemAnnot_2012}
{ISO},
\newblock {\em {ISO 24617-2 Language Resource Management -- Semantic annotation
  framework -- Part2: Dialogue acts}},
\newblock International Organisation for Standardization, Geneva, Switzerland,
  2012.

\bibitem{keizer-bunt-2007-evaluating}
Simon Keizer and Harry Bunt,
\newblock ``Evaluating combinations of dialogue acts for generation,''
\newblock in {\em Proceedings of the \nth{8} SIGdial Workshop on Discourse and
  Dialogue}, Antwerp, Belgium, Sept. 2007, pp. 158--165, Association for
  Computational Linguistics.

\bibitem{henderson-etal-2014-second}
Matthew Henderson, Blaise Thomson, and Jason~D. Williams,
\newblock ``The second dialog state tracking challenge,''
\newblock in {\em Proceedings of the 15th Annual Meeting of the Special
  Interest Group on Discourse and Dialogue ({SIGDIAL})}, Philadelphia, PA,
  U.S.A., June 2014, pp. 263--272, Association for Computational Linguistics.

\bibitem{schatzmann-etal-2007-agenda}
Jost Schatzmann, Blaise Thomson, Karl Weilhammer, Hui Ye, and Steve Young,
\newblock ``Agenda-based user simulation for bootstrapping a {POMDP} dialogue
  system,''
\newblock in {\em Human Language Technologies 2007: The Conference of the North
  {A}merican Chapter of the Association for Computational Linguistics;
  Companion Volume, Short Papers}, Rochester, New York, Apr. 2007, pp.
  149--152, Association for Computational Linguistics.

\bibitem{Sutton1998}
Richard~S. Sutton and Andrew~G. Barto,
\newblock {\em Reinforcement Learning: An Introduction},
\newblock The MIT Press, second edition, 2018.

\bibitem{stoyanchev-etal-icassp2021}
Svetlana Stoyanchev, Simon Keizer, and Rama Doddipatla,
\newblock ``Action state update approach to dialogue management,''
\newblock in {\em ICASSP 2021 - 2021 IEEE International Conference on
  Acoustics, Speech and Signal Processing (ICASSP)}, 2021, pp. 7398--7402.

\bibitem{schatzmann-etal-2009-agenda}
Jost Schatzmann and Steve Young,
\newblock ``The hidden agenda user simulation model,''
\newblock {\em IEEE Transactions on Audio, Speech, and Language Processing},
  vol. 17, no. 4, pp. 733--747, 2009.

\bibitem{kreyssig-etal-2019}
Florian Kreyssig, Inigo Casanueva, Pawel Budzianowski, and Milica Gasic,
\newblock ``Neural user simulation for corpus-based policy optimisation for
  spoken dialogue systems,''
\newblock in {\em Proceedings of the \nth{19} Annual Meeting of the Special
  Interest Group on Discourse and Dialogue}, Melbourne, Australia, 2018.

\bibitem{gasic-etal-asru2011}
Milica Gašić, Filip Jurčíček, Blaise Thomson, Kai Yu, and Steve Young,
\newblock ``On-line policy optimisation of spoken dialogue systems via live
  interaction with human subjects,''
\newblock in {\em IEEE Workshop on Automatic Speech Recognition Understanding
  (ASRU)}, 2011, pp. 312--317.

\bibitem{gasic-etal-icassp2013}
M.~Gašić, C.~Breslin, M.~Henderson, D.~Kim, M.~Szummer, B.~Thomson,
  P.~Tsiakoulis, and S.~Young,
\newblock ``On-line policy optimisation of bayesian spoken dialogue systems via
  human interaction,''
\newblock in {\em IEEE International Conference on Acoustics, Speech and Signal
  Processing (ICASSP)}, 2013, pp. 8367--8371.

\bibitem{Mnih-etal-nature}
V.~Mnih, K.~Kavukcuoglu, D.~Silver, and et~al.,
\newblock ``Human-level control through deep reinforcement learning,''
\newblock {\em Nature}, vol. 518, pp. 529–533, 2015.

\bibitem{Chen_etal-TASLP2019}
Lu~Chen, Zhi Chen, Bowen Tan, Sishan Long, Milica Gašić, and Kai Yu,
\newblock ``Agentgraph: Toward universal dialogue management with structured
  deep reinforcement learning,''
\newblock {\em IEEE/ACM Transactions on Audio, Speech, and Language
  Processing}, vol. 27, no. 9, pp. 1378--1391, 2019.

\bibitem{bunt-2006-dimensions}
Harry Bunt,
\newblock ``Dimensions in dialogue act annotation,''
\newblock in {\em Proceedings of the Fifth International Conference on Language
  Resources and Evaluation ({LREC}{'}06)}, Genoa, Italy, May 2006, European
  Language Resources Association (ELRA).

\end{thebibliography}

\end{document}